\documentclass[journal]{IEEEtran}

\usepackage{cite}

\ifCLASSINFOpdf
 \usepackage{graphicx}
 \usepackage{float}
 \usepackage{placeins}
 \usepackage{multicol}
 \usepackage{lipsum}
 \usepackage{amssymb}
 \usepackage{multirow}
 \usepackage[table,xcdraw]{xcolor}
\usepackage[hidelinks]{hyperref}
\hypersetup{
  colorlinks   = true, 
  urlcolor     = blue, 
  linkcolor    = blue, 
  citecolor   = red 
}
\else
\fi
\usepackage{amsmath}
\usepackage{balance}
\usepackage{hyperref}
\usepackage{breakurl}

\usepackage{rotating}
\usepackage{afterpage}

\begin{document}
%
\title{Artificial Intelligence Methods in In-Cabin Use Cases: A Survey}

\author{
    \IEEEauthorblockN{Yao Rong\IEEEauthorrefmark{1}, Chao Han\IEEEauthorrefmark{2}, Christian Hellert\IEEEauthorrefmark{2}, Antje Loyal\IEEEauthorrefmark{2}, Enkelejda Kasneci\IEEEauthorrefmark{1}}\\
    \IEEEauthorblockA{\IEEEauthorrefmark{1}Human-Computer Interaction, University of T{\"u}bingen, Germany
    \\\{yao.rong, enkelejda.kasneci\}@uni-tuebingen.de}\\
    
    \IEEEauthorblockA{\IEEEauthorrefmark{2}Continental Automotive GmbH, Germany
    \\\{chao.han, christian.hellert, antje.loyal\}@continental-corporation.com}
}

\maketitle

\begin{abstract}
As interest in autonomous driving increases, efforts are being made to meet requirements for the high-level automation of vehicles. In this context, the functionality inside the vehicle cabin plays a key role in ensuring a safe and pleasant journey for driver and passenger alike. At the same time, recent advances in the field of artificial intelligence (AI) have enabled a whole range of new applications and assistance systems to solve automated problems in the vehicle cabin. This paper presents a  thorough survey on existing work that utilizes AI methods for use-cases inside the driving cabin, focusing, in particular, on application scenarios related to (1) driving safety and (2) driving comfort. Results from the surveyed works show that AI technology has a promising future in tackling in-cabin tasks within the autonomous driving aspect.
\end{abstract}


\section{Introduction}
\begin{table*}[t]
\centering
\caption{Use-cases for driving safety.\\ An overview of use-cases for \textbf{Safety}. For each use-case we summarize the typically employed input features and the AI methodology. Round brackets mark the source of the feature: D = Driver, V = Vehicle, O = Outside/Road view.}
\label{tab_safety}
\resizebox{0.9\textwidth}{!}{
\begin{tabular}
{|l |l |l |l |l |}
\hline
\multicolumn{1}{|c|}{\textbf{}}  & { \textbf{Use-case}}                         & { \textbf{Feature}}              & { \textbf{Method}}            & { \textbf{Reference}}                               \\\hline

& \textit{\textbf{Emotion detection}}   & \begin{tabular}[c]{@{}l@{}} physiological information (D)\\ acoustic signals (D) \\ image of driver (D)\end{tabular} & \begin{tabular}[c]{@{}l@{}}FFNN, CNN\\ SVM\\ Fuzzy Logic System \\ GMM (regression model) \end{tabular} &
\begin{tabular}[c]{@{}l@{}} \cite{fan2010using, katsis2008toward,ali2016cnn, wang2008recognition,lisetti2005affective, affectiva, tawari2010speech, al2011novel}\end{tabular} \\
\cline{2-5} 
\textbf{\begin{tabular}[c]{@{}l@{}} Driver \\ \\ Status\\ \\ Monitoring\end{tabular}}
& \textit{\textbf{Fatigue detection}}              & \begin{tabular}[c]{@{}l@{}}eyelid movement (D)\\ mouth movement (D)\\ head posture (D)\\ physiological information (D)\\ vehicle dynamics (V)\end{tabular} & \begin{tabular}[c]{@{}l@{}}FFNN\\ SVM\\ Fuzzy Logic System\end{tabular}      &
\begin{tabular}[c]{@{}l@{}}\cite{azim2014fully, mandal2016towards}\\ \cite{de2017detection, wang2016driver} \end{tabular}                         \\  \cline{2-5}  & \textit{\textbf{Distraction detection}}         & \begin{tabular}[c]{@{}l@{}}image of driver and road (D\&O)\\ physiological information (D) \\ head posture (D) \\ vehicle dynamics (V)\\ driver behavior (D)\end{tabular}                           & \begin{tabular}[c]{@{}l@{}}Semi-Supervised Learning\\ SVM, Random Forests\\ Maximal Information Coefficient \\ CNN \\ GMM (preprocessing)\end{tabular} &
\begin{tabular}[c]{@{}l@{}} \cite{vicente2015driver, li2015predicting, liang2007real, liu2015driver, xing2018driver, abouelnaga2017real, kose2019real, xing2019driver activity,xing2017identification }\end{tabular} \\ \cline{2-5} 
\multirow{-3}{*}{}              & \textit{\textbf{Attention detection}}           & \begin{tabular}[c]{@{}l@{}}eye gaze (D)\\ head posture (D)\\ full images of driver (D)\end{tabular}  & 3D CNN     &
\begin{tabular}[c]{@{}l@{}} \cite{fridman2015driver, borghi2017poseidon, palazzi2018predicting}\end{tabular}     \\ \hline
\multicolumn{1}{|c|}{\textbf{\begin{tabular}[c]{@{}c@{}}\\ Driving\\\end{tabular}}}   & \textit{\textbf{Driver intention analysis}}     & \begin{tabular}[c]{@{}l@{}}vehicle position and dynamics (V)\\ image of driver and road (D\&O) \\head posture (D)\end{tabular}    & \begin{tabular}[c]{@{}l@{}}SVM, Random Forest\\ GMM (regression model)\\ RNN/LSTM\\ HMM \\ 3D CNN\end{tabular} &
\begin{tabular}[c]{@{}l@{}} \cite{kumar2013learning, jain2016brain4cars, jain2016recurrent, mccall2007lane, pugeault2015much, butakov2014personalized, gebert2019end, rong2020driver}\end{tabular} \\ 
\cline{2-5} 
\multicolumn{1}{|c|}{\multirow{-2}{*}{}\textbf{\begin{tabular}[c]{@{}c@{}}Assistance\\\end{tabular}}} & \textit{\textbf{Traffic hazards warning}}     & \begin{tabular}[c]{@{}l@{}}head posture (D)\\ vehicle dynamics (V)\\ image of road (O)\end{tabular}                                                 & Fuzzy logic system       &
\begin{tabular}[c]{@{}l@{}} \cite{rezaei2014look} \end{tabular}   \\ \hline
\textbf{\begin{tabular}[c]{@{}l@{}}Takeover\\ \\ Readiness\end{tabular}}           & \textit{\textbf{Takeover readiness evaluation}} & \begin{tabular}[c]{@{}l@{}}vehicle dynamics (V)\\ eye gaze (D)\\ driver behavior (D)\end{tabular}           & SVM, KNN  &
\begin{tabular}[c]{@{}l@{}} \cite{braunagel2017ready}\end{tabular}                                            \\ \hline
\end{tabular}
}

\end{table*}

Autonomous driving is among the most widely discussed topics in the recent decade. As a new transportation technology, the autonomous vehicle is designed to surpass human drivers in many aspects, particular in safety. However, in order to realize fully autonomous driving, different levels of autonomy are planned to be achieved successively. According to the \textit{Taxonomy and Definitions for Terms Related to On-Road Motor Vehicle Automated Driving Systems} by the SAE International \cite{SAE levels}, there are six different levels leading up to full autonomy, with Level 0 representing ``fully manual" and Level 5 representing ``fully autonomous" driving. The vehicles functioning between Level 0 and Level 5 are all regarded as semi-autonomous vehicles.

Current research and product development is mainly targeting Level 3 (L3) and Level 4 (L4). For L3, the presence of the driver is required to resolve driving situations that are not manageable by automation. The task for autonomous vehicles is to handle driving under certain conditions, such as driving on a highway or in a city traffic jam. Many vehicle manufacturers are now focusing on incorporating L3 automated systems into their products, e.g. Audi \textit{Traffic Jam Pilot} \cite{Audi}. From L4 on, requests of the takeover from a human driver are no longer necessary. The vehicle is required to analyze driving situations and make informed decisions, like when to change lanes, turn, accelerate, or brake. Even in the case of a device failure, the autonomous systems should be able to safely handle these actions independently. In L4, however, manual intervention does remain for particularly challenging circumstances, such as a system failure. L5 vehicles can operate under all situations, but also provide a more refined, higher quality of services.

Currently, the autonomous vehicles have achieved L3 and progressing towards L4. Human drivers are still the main decision makers and supervise the entire system. Consequently, an aspect of ongoing research in L3 is to find the optimal way of assisting human drivers and to provide a smooth and safe transition from human to autonomous driving and back again. Driver-related activities within the vehicle's cabin should be monitored and analyzed by the system to achieve not only a safe and comfortable drive, but to ensure the system's ability to smoothly handle a takeover situation.
Most of the tasks in autonomous driving are related to ``\textit{perception}". As human beings, we receive information mostly through vision and speech. We analyze this information and respond accordingly to different events. To endow vehicles with the same capability of understanding, researchers are mounting AI technology on autonomous vehicles to automate the perception of surroundings. Additionally, with emerging technologies such as Augmented/Virtual reality (AR/VR) new ways of personalized driving assistance,  information, navigation and entertainment \cite{Michigan, civilmaps, wayray} have been enabled. 
Given the broad application of AI technologies inside the driving cabin, we are performing a thorough survey of existing studies conducted by researchers and system developers. Our motivation is to identify and highlight the similarities and differences within existing works in order to envision new applications. In this context, similarities means the identification of different applications using the same input data modality or algorithms. This often indicates an emerging trend of research. Moreover, resource efficiency can be improved if one input feature (or hardware) can be used in different applications. A diversity of algorithms can be used to solve similar problems. Reviewing and referring to existing works can serve as an inspiration for readers seeking concrete solutions for specific tasks in autonomous driving. We aim to provide a clear overview of commonly used hardware and algorithms, with a strong focus on the SAEs L3 and L4. L3 and L4 will be discussed with an emphasis on considerations for safety and comfort.

The paper is organized following: in Section \ref{sec: safety}, we will discuss different applications that contribute to safety and employed methodological approaches. Section \ref{sec: comfort} introduces tasks aimed at comfortable driving. In the last section, we summarize all the surveyed works and provide a brief outlook. 

\section{In-cabin use-cases for driving safety}
\label{sec: safety}
According to the \textit{National Highway Traffic Safety Administration} (NHTSA) in the USA, $94\%$ of serious accidents are caused by errors behind the wheel \cite{NHTSA}. One important task for an autonomous driving system is to ensure the safety of the driver, passengers and other vehicles and pedestrians on the road. Since SAE L3 and L4 require a driver’s presence, the system is responsible for monitoring the driver. For instance, the system needs to assess whether or not the driver is in a proper state for driving and to assist the driver in the decision-making process. Table \ref{tab_safety} presents a short overview of the use-cases discussed in the following sections.

\subsection{Driver status monitoring}
\label{sec:driver monitoring}
Various Driver Monitoring Systems (DMS) have been developed over the past few years. Due to the rapid development of AI technology, some mature systems are currently being utilized in the market. For example, Seeing Machines \cite{seeingmachines}, Valeo Driver Monitoring \cite{valeo}, and SmartEye Driver Monitoring System \cite{smatereye}. These systems are usually based on image information from a camera mounted in front of the driver. They infer information about the driver based on the analysis of her facial expression, eye gazes or head posture. Physiological signals, such as heart rate and skin temperature can also contain valuable information about the driver. Utilizing this information is helpful when gauging the driver's vigilance, emotions and level of attention or distraction. 

\subsubsection{Emotion detection}
The emotional status of a driver can heavily influence the decision-making process and overall behavior on the road. It is important to analyze the emotional status of the driver and process this information accordingly within the automated system. In particular, ``Aggressive driving," as defined by NHTSA, has been researched for decades regarding its negative influence on road safety \cite{aggressive driving}. Drivers often respond to aggressive acts by another driver with anger and mirrored aggression. \cite{road rage}. Due to this common response from drivers, it is important to monitor driver emotions periodically. Automated recognition of a driver's emotional state can capture warnings of aggressive or distracted driving due to
``road rage" before behavior escalates, resulting in a safer driving experience.

Since emotions correlate strongly with facial expressions, automated methods for emotion recognition based on images have been the focus of research over the last two decades. A few of these approaches, \cite{shan2005robust, bartlett2003real} use Cohn and Kanade's dataset\cite{kanade2000comprehensive}, which contains a large number of facial image sequences from different people. \cite{bartlett2003real} proposes a system that first locates the face in the image and then classifies the emotions based on the Gabor magnitude representations of the located faces. An approach with AdaSVM provides the best performance in this work: Gabor features chosen by Adaboost were used as the training input for Supported Vector Machine (SVM) classifier. \cite{shan2005robust} uses Local Binary Patterns (LBP) as the discriminative features rather than Gabor features which allows for a very fast feature extraction. Similarly, an SVM is employed as the classifier for emotion recognition.

When it comes to in-cabin driver emotion detection, image, speech and physiological signals are often used for detecting emotion. A motion estimation system named ``Affectiva" \cite{affectiva}, which is also applied in automotive applications, uses driver facial images and speech signals. Most of the research work on this topic focuses on physiological signals due to suitability and accuracy. In \cite{ali2016cnn, katsis2008toward, fan2010using}, biopotentials are measured by various medical techniques: electromyogram (EMG) in \cite{katsis2008toward}, electrocardiogram (ECG) in \cite{ali2016cnn, katsis2008toward}, electroencephalogram (EEG) in \cite{fan2010using} and electroencephalogram activity (EDA) in \cite{ali2016cnn, katsis2008toward}. Besides biopotential, skin temperature is also used in \cite{ali2016cnn, wang2008recognition,lisetti2005affective}, as is respiration in \cite{katsis2008toward, wang2008recognition} and heart rate in \cite{lisetti2005affective}. In addition to physiological features, a driver's acoustic signals are processed for this same purpose in \cite{tawari2010speech, al2011novel}. Speech may not offer as robust a result as biosignals, but acquisition of acoustic signals is simple and unobtrusive.

A diverse selection of emotions is required to effectively train machine learning models. Over the last seven years, a massive amount of video and audio data from all over the world has been collected for the emotion AI system \cite{affectiva}. In \cite{lisetti2005affective, wang2008recognition, katsis2008toward}, data is recorded when different affective behaviors from drivers are elicited in simulated driving scenarios in the lab. In \cite{tawari2010speech}, however, real world speech clips are used. A publicly available speech database called Emo-DB \cite{burkhardt2005database} is used in \cite{al2011novel}.

With the help of large amounts of real world data, very deep Convolutional Neural Networks (CNNs) are trained for the classification of seven different emotion \cite{affectiva}. In \cite{ali2016cnn, katsis2008toward}, four different classes of emotion (excited, relaxed, angry and sad) are detectable by Feed Forward Neural Networks (FFNNs). \cite{ali2016cnn} uses cellular neural network, while \cite{katsis2008toward} combines FFNN and fuzzy inference systems. \cite{lisetti2005affective} also uses FFNNs but trains with different optimizers: Marquardt Backpropagation (MBP) and Resilient
Backpropagation (RBP) algorithms. The best results are achieved by RBP amongst five different emotional states with 91.9\% accuracy. The authors from \cite{wang2008recognition} propose a novel latent variable model and also introduce the temporal state into the model. Training this model is similar to training a Gaussian Mixture model (GMM). 

Audio streams are used as system inputs and extract acoustic features like speech intensity, pitch, and Mel-Frequency Cepstral Coefficients (MFCC) in \cite{tawari2010speech, al2011novel}. An SVM and Bayesian Quadratic Discriminate Classifier are trained in \cite{tawari2010speech} and \cite{al2011novel}, respectively. Moreover, \cite{tawari2010speech} uses speech enhancement to resist the influence of noisy background influence. It also shows that including gender information results in better overall recognition. 

\subsubsection{Fatigue detection}
Drowsy driving greatly impacts the safety of those on the road. It is necessary to remind drivers to rest when the fatigue is detected. The most popular feature for measuring fatigue is eyelid movement, particularly the percentage of eyelid closure (PERCLOS)\cite{wierwille1994research}. Other useful information can include facial expressions, physiological information (heart rate) and vehicle data (car speed, steering wheel angle, position on the lane).

For successful fatigue detection, eye metrics are useful \cite{azim2014fully, mandal2016towards, wang2016driver, de2017detection, baccour19camera, baccour20camera}. Such features can be collected simply by using a regular camera mounted in front of the driver. In \cite{azim2014fully}, yawning (mouth movement) is measured along with eye closure; in \cite{wang2016driver}, vehicle data is also proven to be useful. \cite{baccour19camera} uses the velocity of eyelid to detect the eye blink for assessing drowsiness. Eye blinking and head movements are used together as input signals of logic regression models for drowsiness state classification in \cite{baccour20camera}. \cite{de2017detection} compares the detection accuracy using behavioral data (eye and head movement), physiological information and vehicle data.

Different machine learning models can be applied to determine whether or not the driver is exhausted. \cite{azim2014fully} uses a Fuzzy Expert System to classify the state of the driver, while \cite{mandal2016towards} deploys a binary SVM classifier for detecting open and closed eyes. \cite{de2017detection, wang2016driver} show that the FFNN is also suitable for measuring levels of drowsiness. Especially in \cite{de2017detection}, the FFNN can even predict when the driver will reach a given level. 

\subsubsection{Distraction detection}
Distraction is another major threat to driving safety, motivating researchers to study activities that often lead to preoccupied driving.

According to \cite{ranney2001nhtsa}, distraction has four distinct categories: visual, cognitive, auditory, and bio-mechanical. Visual distraction is defined as ``eye-off-the-road", which is obvious to detect. In this instance, eye gaze is an essential feature for detection. In \cite{vicente2015driver}, the proposed method estimates a 3D head pose and a 3D eye gaze direction using a low-cost CCD (charge-coupled device) camera. Estimations are measured with respect to the camera coordinate system. With the rotation matrix from the camera coordinate to the world coordinate system, the driver's observance of the road can be measured. An SVM classifier is used first for detecting sunglasses. If sunglasses are detected, the estimation relies only on the head pose. \cite{tian2019standardized} proposes a standardized framework for evaluating a system, which tracks driver head movements to alert in case the driver is distracted. Such a standard makes it possible to fairly evaluate different driver head tracking systems. In addition, this framework introduces a ground-truth data acquisition system, PolhemusTM Patriot, and takes driver-related information (gender, race and age, etc.) into account. \cite{liang2007real} uses eye movements and driving data to classify normal and distracted driving in real time. It also proves that the SVM classifier is suitable for such a task.

Compared with visual distraction, cognitive distraction such as daydreaming or becoming ``lost in thought" is harder to detect. Cognitive distraction is also called ``mind-off-the-road", indicating a loss of situation awareness. Facial expressions and driving performance reflect this distraction. \cite{li2015predicting} explores the effect of both distractions with the help of multi-modality features from CAN-Bus, microphone, and camera recording road and driver. Classifiers employ these feature representations to discriminate between different distraction levels. The causes of cognitive distractions are variable. Estimation of drivers’ workloads can also impact the cognitive state of the driver. To measure workloads, \cite{xing2018driver} proposes a new nonlinear causality detection method called error reduction ratio causality, which identifies the important variables. The variables used here include Skin Conductance Response (SCR), hand temperature and heart rate, as well as GPS position and acceleration recorded from real-world driving. An SVM is trained afterwards to select the right model for measurement.

\cite{liu2015driver} studies audio-cognitive distraction. The task for the driver is to count how many times each of the target sounds appear. An eye tracker records eye and head movement data. This data is then used to train a Laplacian SVM and Semi-Supervised Extreme Learning Machine. The study also proves that using a semi-supervised learning algorithm outperforms supervised learning when giving more unlabeled data.

Bio-mechanical distraction refers to adjusting devices manually. For example, adjusting the radio. The solution is to simplify the Human-Machine-Interface (HMI) in the cabin, which will be discussed in Section \ref{sec: comfort}.

Performing secondary tasks always causes more than one distraction. Distracting secondary tasks include talking on a cell phone or drinking/eating. Deep neural networks can recognize these behaviors which are very helpful in action recognition. For example in \cite{xing2019driver activity,xing2017identification}, seven activities are divided into two groups: normal driving (normal driving, right mirror checking, rear mirror checking and left mirror checking) and distraction (using in-vehicle radio device, texting and  answering the mobile phone). The dataset is collected using Kinect, so the images and the coordinates of head centre or upper body joints are recorded. \cite{xing2017identification} uses Random Forests (RF), Maximal Information Coefficient (MIC) and a FFNN as classifiers using the head and body features. \cite{xing2019driver activity} only uses images of drivers. The images are first processed by a GMM to segment the driver's body, and then used for CNNs training. The CNN backbones used in experiments are AlexNet, GoogLeNet, and ResNet50. The best performance is achieved by AlexNet which also surpasses the result in \cite{xing2017identification}. In \cite{abouelnaga2017real, kose2019real}, CNNs such as AlexNet, InceptionV3 and BN-Inception are trained in end-to-end manner. These networks achieve distracting activity recognition with high accuracy.

\subsubsection{Attention detection}
Another important task for DMS is to understand where the driver is looking while driving. With a high-level criticallity of the event detected (e.g. a pedestrian crossing the street), the system warns the driver if the driver is not paying attention~\cite{kasneci2015online,kasneci2017aggregating}. This task is one specific use-case in visual attention modeling. Visual saliency and gaze are common tools for measuring the attentive area.

Eye-tracking glasses have the ability to track the precious position of the gaze, but it is challenging for the driver to wear equipment while driving. In this case, head posture estimation assists with gaze estimation. In \cite{fridman2015driver}, a pipeline is proposed: facial feature detection and tracking -- (3D) head posture estimation -- gaze region estimation. Besides using handcrafted features such as facial landmarks, \cite{borghi2017poseidon} proposes a deep CNN for localizing the driver's head and shoulder position in the depth images.

It is also possible to predict the focus of attention without using head posture information. For instance, in \cite{palazzi2018predicting}, the raw video, optical flow and semantic segmentation information are fed to a multi-branch 3D-CNN for end-to-end training, in order to predict the focus area on the road image. In the future, attention prediction for human drivers can contribute to attention mechanisms for autonomous perception functions.

\subsection{Driving assistance}
In Section \ref{sec:driver monitoring}, we discussed the Driver Monitoring System, a system that focuses on and contributes to safe driving. The Advanced Driver Assistance System (ADAS) is also designed to avoid accidents by alerting the driver to potential problems or by taking over the control of the vehicle. In the last decades, functions such as anticipating the intention of drivers and analyzing on-road traffic have also been studied. This section introduces these functions integrated into ADAS.

\subsubsection{Driver intention analysis}
Accelerating, braking, steering, turning and lane changing are common tasks during driving. Wrong decisions can result in critical situations or triggering accidents. ADAS assists with lane keeping or changing and prevents some dangerous maneuvers. In order to assist the driver, it has to understand the driving context. \cite{pugeault2015much} uses visual gist as the image descriptor for pre-attentive perception. The images are captured by three on-board cameras. A Random Forest (RF) classifier trained with the gist features can differentiate road contexts such as single-lane, crossing, or T-junction. Furthermore, it can successfully predict driving actions in real time using driving context information. 

An important driving behavior is lane changing. In \cite{kumar2013learning, butakov2014personalized, jain2016brain4cars, jain2016recurrent, mccall2007lane}, lane changing behavior is anticipated. \cite{kumar2013learning} predicts three classes: right/left lane change and no lane change. The features are collected by a vision and Inertial Measurement Unit (IMU) based lane tracker. The position of the vehicle in respect to the lane, more specifically the lateral position and the steering angle, are recorded. The proposed prediction model includes a Bayesian filter and an SVM classifier. The Bayesian filter takes the output from the SVM and produces a final prediction. \cite{mccall2007lane} predicts whether or not lane changing occurs with the help of the Sparse Bayesian Learning (SBL) model. The input features are lane positional information acquired from the camera focused on the road, vehicle parameters from CAN-Bus, and driver head posture obtained from the image of the driver. In \cite{jain2016brain4cars, jain2016recurrent}, more driving behaviors are included in addition to the three lane changing classes, i.e. right/left turn. The input information sources are various in this dataset. They include videos of drivers and the road outside the vehicle, vehicle dynamics, GPS, and street maps. \cite{jain2016recurrent} makes use of all this information and trains a Recurrent Neural Network (RNN) with Long-Short Term Memory Cells (LSTM). According to the results in \cite{jain2016brain4cars}, this architecture achieves the best result when compared with SVM, RF or Hidden Markov Model (HMM). Moreover, it anticipates the action with an average 3.58s. Using videos of drivers, end-to-end prediction is also accurate. For instance, in \cite{gebert2019end} the 3D ResNeXt-101 with a LSTM layer on the top is trained in end-to-end style. The results in \cite{rong2020driver} prove that videos towards roads have complementary information as driver videos, which should also be considered in driver maneuver prediction. \cite{butakov2014personalized} takes the personalities of drivers into account because ADAS should comply with the driver's habits to ensure overall safety. It proposes using a GMM to adjust the sinusoidal lane change kinematic model according to individual driving styles.

Finally yet importantly, \cite{xing2019driver} provides an overview of a multi-module Driver Intention Inference (DII) system designed for lane changing intention detection. This system consists of different modules: traffic context perception module, vehicle dynamic module, driver behavior recognition module and driver intention inference module. From this work, we can see an emerging trend of multi-module fusion in ADAS.

\subsubsection{Traffic hazards warning}
Not only should ADAS focus on the intention of the driver, but it should also simultaneously observe on-road traffic. This can prevent some traffic accidents by correlating information and notifying the driver in a timely manner. On-road hazards include rear-end crashes, unnoticed pedestrians, speed breakers or traffic signs.

\begin{figure}[h]
	\centering
	\includegraphics[width=0.85\linewidth]{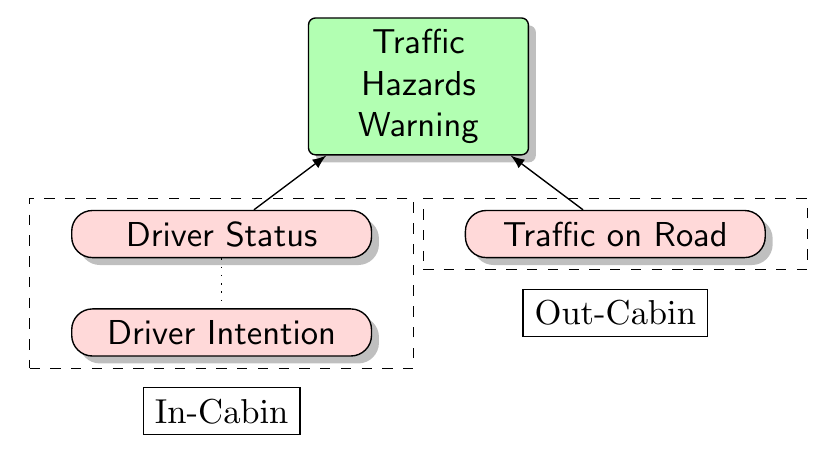}
	\caption{Traffic hazards warning system includes both in- and out-cabin analysis. }
	\label{fig: traffic hazards warning}
\end{figure}

One possible solution for this task is to combine the \textit{driver intention prediction/driver status detection} with \textit{on-road traffic detection}. It requires driver monitoring, object detection/tracking, and data fusion modules to work simultaneously. Fig. \ref{fig: traffic hazards warning} shows the components of the system. Traffic detection that only uses on-road information is not related to in-cabin applications and will not be discussed.

The system in \cite{rezaei2014look} consists of two modules in Fig. \ref{fig: traffic hazards warning}. Driver head posture estimation is a preliminary part of driver attention analysis. A 3D face model is trained using an asymmetric face appearance model. Mapping 2D feature-points into a 3D face helps to determine the direction of the driver's attention. The second component of the driver-assistance system is road traffic detection, which uses global Haar-like features (GHaar) classifier to detect vehicles ahead on the road. Additionally, the system can estimate the distance and the angle between the detected vehicle and the ego vehicle in relation to the right lane of the road. A fuzzy logic system extrapolates future driving risks based on driver and on-road information.

Besides other vehicles, pedestrians and bicycles are other important factors on road. In \cite{kim2016look}, the authors developed a pedestrian collision warning system, equipped with a volumetric head-up display (HUD) in the cabin to identify when and where pedestrians are approaching. This work also shows that the Augmented Reality (AR) technique is both effective and intuitive for warning systems within the cabin.

\subsection{Take-over readiness evaluation}
As mentioned, at SAE L3 the human driver should stand by and be prepared to take over control of the vehicle. \textit{Takeover readiness} defines the driver's ability to regain control of the vehicle from the automated mode. Non-driving related tasks during automated driving may interfere with a driver's ability to regain control of the vehicle \cite{tanshi2019modeling}. Thus, it is necessary to help the driver stay prepared for a takeover. In this section, we discuss some methodologies that measure driver takeover readiness.

To study the readiness of the driver, takeover request (TOR) time is a key term. TOR measures the time between the request for takeover and the critical situation (by which time the driver msut maintain control). Determining when to alert the driver to a takeover situation is critical. \cite{kim2017takeover} studies four different TOR times. The results show that the TOR resulting from the performance-based method provides the shortest reaction time and highest satisfaction for drivers. This performance-based method considers the influence of driving behaviors. It was originally designed for the airborne collision avoidance system. Besides the TOR time, there are other factors that may influence takeover behavior. Factors may include traffic situation complexity, ego-motion of the vehicle, and type of secondary tasks, etc. \cite{tanshi2019modeling} studies how the complexity of the driving task and secondary task impact takeover reaction time. A mathematical formula estimates takeover reaction time based on experimental data. \cite{kim2018design} creates a concept system that can estimate readiness directly by using driver behavior information and biometric data. Extracted eye gazes and head movements are driver behaviors while heart rate and respiration rate are considered biometric data.

There is relatively little research employing machine learning methods to estimate driver readiness, with the exception of \cite{braunagel2015driver,braunagel2017online,braunagel2017ready}. The authors use multi-modality data to train different classifiers, such as K-Nearest Neighbors (KNN) and SVM. The studied data includes the maximum deviation from the lane center, the minimum distance to the leading vehicle and drivers' eye gazes and behaviors. These classifiers predict the quality of takeover readiness. The best result is achieved by a linear SVM: the accuracy is 79\%.

In addition to the estimation of takeover readiness, the system is responsible for keeping the driver constantly aware of the situation both inside and outside of the vehicle. An Interactive Automation Control System (IACS) designed in \cite{olaverri2018automated} keeps the driver aware of the TOR on a display. Experiment results show that the response time to TOR and the total number of collisions decreases due to support from this system. \cite{wiegand2018early} proposes a system which employs AR. In this system, AR is used to show a digital twin of the driver's car on in simulation of a potential accident where the TOR is necessary. After alerting the driver to the coming situation, the TOR is executed. This work indicates that a simulation in cockpit can help the driver better understand traffic situations and handle the TOR more effectively. 

One limitation is that all projects presented here are conducted using driving simulators. Since the takeover task is a safety-critical issue, more experiments should be conducted in real-world driving situations.

\section{In-Cabin Use-cases for Driving Comfort}
\label{sec: comfort}
Autonomous vehicle technology makes driving not only safe but also relaxing. Improving driver and passenger comfort level is another key research topic. Tasks in the comfort sector are generally non-driving related tasks. In this section, we introduce some works aiming to optimizing in-cabin operating systems by making vehicles more intelligent.

\subsection{Convenience}
``Convenience" describes the ability of the system to accomplish non-driving related tasks automatically according to the needs of drivers and passengers. An intelligent system should recognize needs in an accurate and timely manner. In order to perceive needs, AI methods are very suitable because they can analyze human actions and the information encoded within human actions. A new dataset named ``Drive\&Act" is collected for driver action recognition purpose \cite{martindrive}. It is collected in driving mode as well as in automated driving mode, and the behaviors are fine-grained labeled. This dataset includes many secondary task actions like \textit{putting on sunglasses} or \textit{reading magazines}. The videos are recorded by six synchronized cameras inside the cabin in RGB, depth, infrared and bodypose modalities. Recognizing these behaviors correctly can increase comfort. For instance, the visor should flip down automatically when the driver is putting on sunglasses. The appearance of this dataset supplements a large benchmark for in-cabin action recognition. The authors in \cite{martindrive} also train different models with this dataset. The best performance is achieved by the 3D CNN-based model. Results indicate that AI methods have a promising future for in-cabin applications.

Listening to music can provide drivers and passengers with a more comfortable journey. Research such as \cite{hu2015safedj} shows that listening to suitable music can improve the driver's mood and fatigue state resulting in improved driving performance. \cite{hu2015safedj} proposes a framework which detects the driver's mood-fatigue status and recommends music accordingly. This framework makes use of different smartphone sensors to gauge each drivers' specific situation and to employ intelligent analysis. For example, the system will engage the closest algorithm to classify different music moods.

\subsection{Human-Machine-Interface}
The more functional automated vehicles are, the more complex HMI can become. Some crucial principles are mentioned in \cite{carsten2019can, benderius2017best} for designing the HMI: HMI should both provide comfort and stimulate an appropriate level of attention from users. HMI should maintain minimal content in order to reduce distraction. For instance, \cite{tian2014study} investigates the position of the display for the haptic rotary device in a conventional vehicle HMI system. The results show that cluster display position reduces lane position deviation during secondary tasks.

The authors in \cite{lindemann2016exploring} propose using AR to realize a multi-layer floating user interface system in the vehicle. This system employs stereoscopic depth to arrange different information on 3-layer displays. Critical information, such as ``low fuel" warning, is shared on the nearest screen. Less critical items are shifted to the back layers and blurred. This system aims at providing a large amount of information without greatly distracting drivers.

Hand gestures and speech are becoming a popular means of simplifying HMI systems because they reduce visual and bio-mechanical distraction during driving. Different sensors and recognition algorithms are used for hand gesture recognition in the vehicular environment. For example, (1) \cite{smith2018gesture} uses mm-wavelength radar sensor and trains a Random Forest. On average, the system performs above 90\% accuracy for all six gestures classes; (2) in \cite{manganaro2019hand}, multiple modalities including RGB, depth/infrared images and 3D hand joints are tested. They train two networks: A C3D network and a Long-Short Term Memory (LSTM). The best model, with a recognition accuracy of 94.4\% for 12 classes, is the LSTM model, using 3D hand joints as input modality. In speech recognition, special uses for driving scenarios are explored. Some examples include: natural language analysis based on a RNN architecture for commands like ``set/change destination or driving speed" in \cite{okur2019natural}, or the the vehicle control system's defense strategy using an SVM classifier that can resist attacks from hidden voice commands in \cite{zhou2019hidden}.

Another traditional HMI element in vehicles is the HVAC (Heating, ventilation, and air conditioning) system. Normally, the controllers are hand-coded, requiring attention from the driver. In \cite{stark2017supervised}, a control system deploying NN architecture can realize automatic control of the cabin's thermal environment. At first, the model collects data while the user is adjusting the system. After training, the model can learn the user's preference and control the thermal environment accordingly. Different machine learning techniques can be used to realize this goal. In \cite{brusey2018reinforcement}, the automatic control is realized using Reinforcement Learning (RL). It should be noted that the RL controller consumes less energy and produces a more comfortable environment than manual control approaches.

For fully autonomous cars, \cite{benderius2017best} proposes that HMI should only contain commands for ``start", ``stop" and ``choose the destination". Additionally, other interfaces for entertainment or maps should be included in personal mobile devices. The advantages are the separation of safety critical functions from non-critical ones, whose personalization remains.

As the SAE level increases, drivers can focus less on driving tasks and have more access to HMIs. Human factors become more influential in the HMI systems. \cite{bengler2020hmi} introduces an HMI framework which clusters human factors (of both drivers and other users of the road) as dynamic factors. Different HMIs are chosen depending on these influential factors. They also propose an external HMI for communicating with other users on the road. One specific and important human factor for autonomous driving is trust in the vehicle. \cite{ekman2017creating} focuses on how to increase human trust for an autonomous car via HMIs. The authors suggest that HMI framework should take multiple events over a period of time into account rather than focus on one isolated event.

\subsection{Navigation}
Navigation is one of the most pronounced functions in modern vehicles. Many drivers have experienced difficulty, trying to concentrate on the road while viewing a personal navigation device. Using an AR Head-Up Display (HUD) to show the navigational path, traffic signs, and landmarks is a practical solution. The work in \cite{medenica2011augmented} proves that drivers prefer navigation using AR HUD to other traditional navigation devices, namely egocentric street view and map view which shows the vehicle within the context of its surroundings on the LCD display. On the HUD, directions are listed on a narrow semi-transparent surface that is suspended above the center of the road at a height of about 2 meters. Moreover, according to eye gazes measurements, drivers spend 5.7 sec and 4.2 sec more per minute looking at the road ahead in comparison to LCD street view and map view, respectively. 
 
In \cite{abdi2015vehicle}, the framework can detect vehicles and traffic signs and project them onto the AR-HUD, helping drivers to avoid some dangerous accidents in the process. For detection, the framework uses AdaBoost learning algorithm to train with the Haar features of vehicles and traffic signs. The next stage after detection is to find positions on the HUD for the projection of virtual objects. To do this, camera parameters and relative position of the camera, with respect to objects, are required for calculation. With the help of AR, virtual objects are attached to real objects. In this case, drivers will be alerted to critical information on the road in an unobtrusive way.

An investigation of the effectiveness of different presentations of AR enhanced navigational instructions in \cite{bolton2015investigation} shows: The most effective arrangement is to use boxes that enclose a landmark, such as ``turn right in 120 meters". The response times and success rates are enhanced by 43.1\% and 26.2\% compared to the conventional representation (only the sign).

\balance
\section{Conclusion}
\begin{figure}[h]
    \centering
   \includegraphics[scale=0.75]{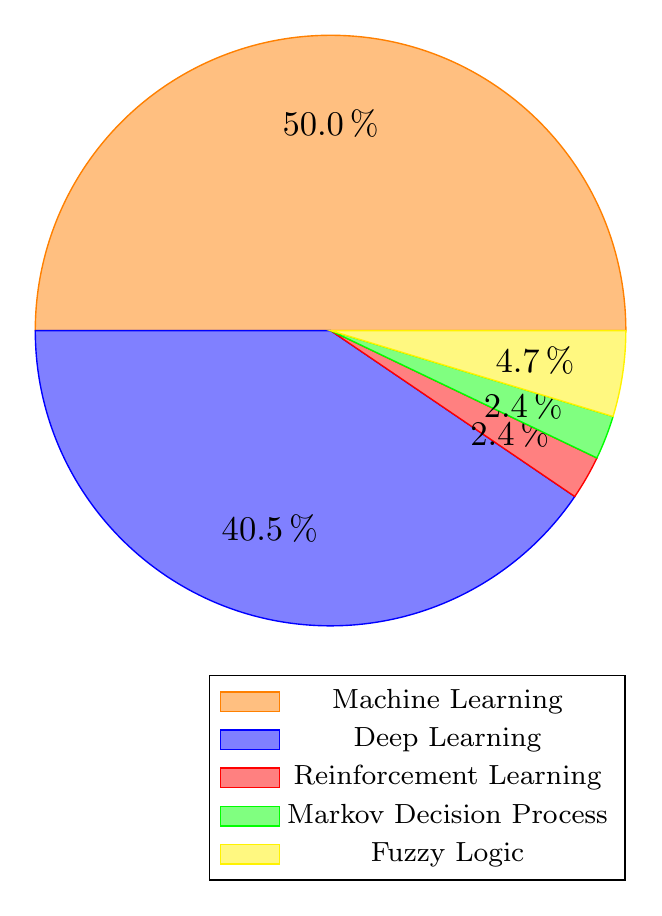}
    \caption{Frequency of utilization of different AI methods in in-cabin use-cases}
    \label{fig:pie chart}
\end{figure}

In this section, we summarize all AI techniques implemented in the in-cabin use-cases we reviewed as well as corresponding features of these applications. 
The in-cabin use-cases can be abstracted into the following topics: classification problem, regression problem and sensor fusion problem. For example, to predict whether or not the driver is tired (in \cite{mandal2016towards}) is a classification problem. To predict the drowsiness level (in \cite{de2017detection}) is a regression problem. A typical occasion for sensor fusion is the ``traffic hazards warning" system proposed in \cite{rezaei2014look}. When enough data is provided, AI methods can easily tackle the three problems outlined. It also explains the frequency of utilization of five techniques shown in Fig. \ref{fig:pie chart}.

Fig. \ref{fig:pie chart} shows the different AI techniques in all examined papers. It is worth noting that ``Deep Learning" refers to the learning algorithms that use layered structures (Artificial Neural Networks). Although it is a subset of ``Machine Learning", it is regarded as a separated set due to its importance in Computer Vision research. The ``Machine Learning" set refers to algorithms with the exception of ``Deep Learning".
There are a total 42 works that utilize AI techniques. Since Machine Learning algorithms and Deep Learning networks are very effective when solving classification and regression problems, both dominate the surveyed works. Concretely, 50.0\% (21 papers) of the applications were solved by Machine Learning algorithms and 40.5\% (17 papers) used Deep Learning networks. The Fuzzy Logic System (4.7\%) is used when there are multiple inputs from different sensors, as exhibited in \cite{azim2014fully, kim2016look}.

We summarize the use-cases discused in this paper and their relationship to SAE L3, L4 and L5 in the Table \ref{sae table}. The \checkmark indicates that the use-case is an important function at this level. As shown in the Table \ref{sae table}, driving assistance, takeover readiness and navigation are no longer necessary in L4 and L5 because a human driver will not intervene. The purpose of driver status monitoring also changes from L3 to L4. The L3 system focuses on driver anomaly while L4 and L5 are concerned with passenger emotion and satisfaction.

\begin{table}[!htb]
\caption{Use-cases and their implementations in different SAE levels (from L3 to L5)}
\label{sae table}
\centering
\begin{tabular}{|l|l|l|l|}
\hline& L3 & L4 & L5 \\ \hline
\begin{tabular}[c]{@{}l@{}}driver status monitoring\end{tabular} &  \checkmark  &  \checkmark  &  \checkmark  \\ \hline
\begin{tabular}[c]{@{}l@{}}driving assistance\end{tabular}  &  \checkmark  &    &    \\ \hline
\begin{tabular}[c]{@{}l@{}}takeover readiness\end{tabular}  & \checkmark   &    &    \\ \hline
convenience  & \checkmark   &   \checkmark &   \checkmark \\ \hline
HMI          &  \checkmark  &  \checkmark  &  \checkmark  \\ \hline
navigation    &  \checkmark  &    &    \\ \hline
\end{tabular}
\end{table}

At the end of this paper, Table \ref{hardware table} itemizes the different hardware employed for data acquisition in examined works utilizing AI techniques. The hardware is categorized into eight different types, as shown in the first column. In the second column, the hardware names or models are listed. Some are marked with ``unknown" when the name is not mentioned in the original work. 

\begin{figure*}[h]
  \includegraphics[width=0.95\linewidth]{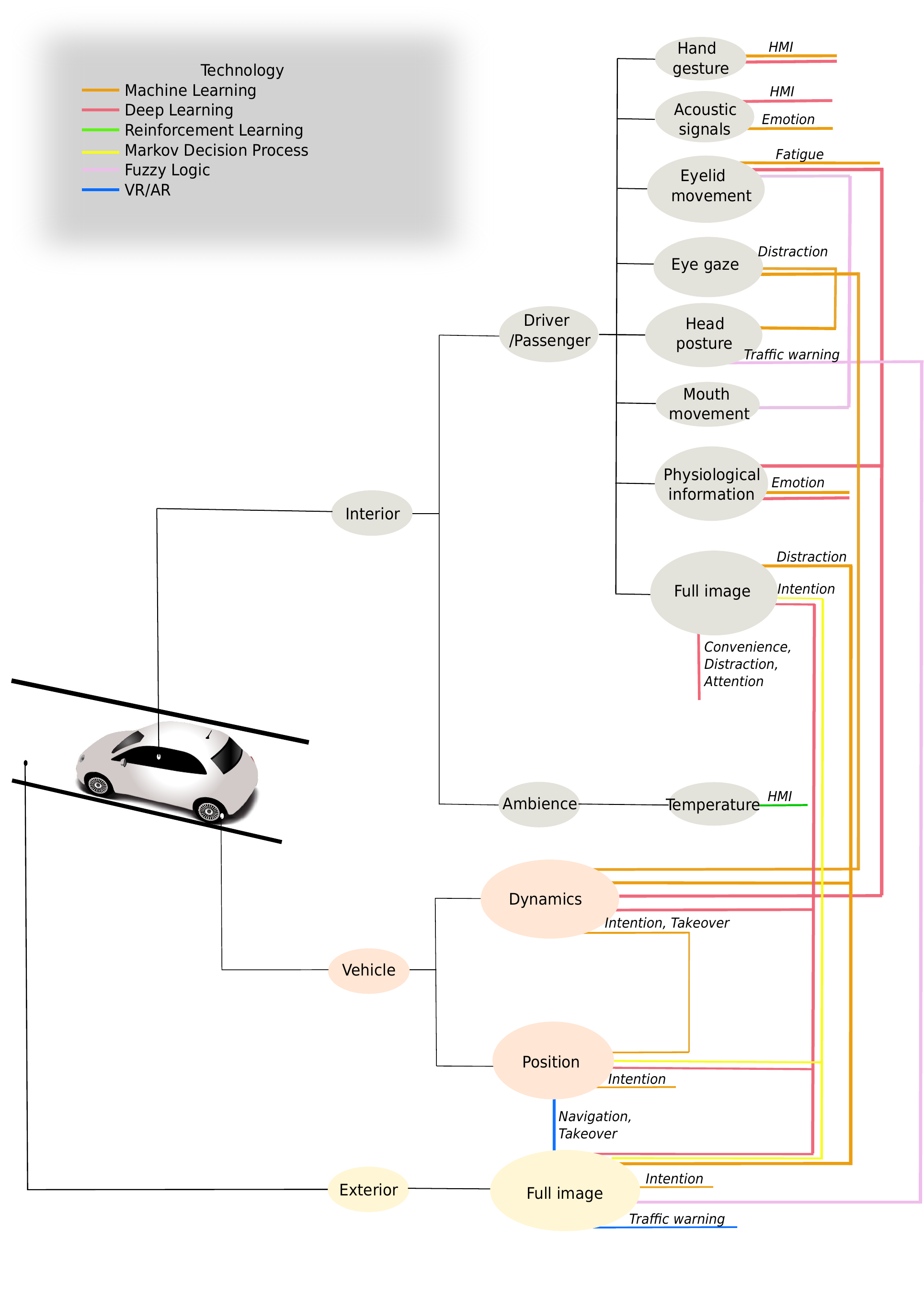}
  \caption{Summary of concrete use-cases, their techniques and features.}
  \label{fig: overview}
\end{figure*}

Fig \ref{fig: overview} summarizes all the use-cases examined in our survey. The features are depicted as ``leaves" in a tree structure. Lines of different colors represent different techniques. The use-cases are described in keywords. If a line emerges from the leaf with an open ending, this means that the application only uses this feature as the input. Typically, the applications use more than one feature, marked with connections in the figure. The use-case is the nearest keyword above the line (or on the right side of the line). From this overview, the following is apparent for in-cabin use-cases: (1) important input features are a driver's eye and head movement, full images of drivers/roads and vehicle position and dynamics; (2) popular techniques are Machine Learning and Deep Learning; (3) research focuses are distraction detection, HMI design, and driver intention analysis.

Fig. \ref{fig: overview} also includes features that are used in different applications. For example, the image of the driver is used widely in distraction and intention detection, as well as for convenience purposes. For the future work, a high-level module integrated with different functionalities will be considered. This module should have a manager that can coordinate the work of different sub-modules. In this way, the resource of the vehicle is saved and different modules can support one another to achieve a holistic solution.




\bibliographystyle{IEEEtran}

\afterpage{%
    \clearpage

\begin{sidewaystable*}[]
\caption{Overview of the hardware utilization for data acquisition in examined papers}
\label{hardware table}
\resizebox{\textwidth}{!}{
\centering
\begin{tabular}{|c|l|l|l|l|l|}
\hline
\begin{tabular}[c]{@{}c@{}}\textbf{Hardware}\\\textbf{ Category}\end{tabular}     & \multicolumn{1}{c|}{\textbf{Hardware}}  & \multicolumn{1}{c|}{\textbf{Feature}}  & \multicolumn{1}{c|}{\textbf{Specifications}} &
\multicolumn{1}{c|}{\textbf{Dataset}} &
\multicolumn{1}{c|}{\textbf{Reference}} \\ \hline
\multirow{15}{*}{camera} & \begin{tabular}[c]{@{}l@{}}Near infrared (NIR) \\ charged coupled device (CCD) \\ camera\end{tabular}             & \begin{tabular}[c]{@{}l@{}}eyelid movement\\ from facial images\end{tabular}             & IR; 25 fps     & test on 3 videos, 100 frames per video & \cite{azim2014fully}     \\ \cline{2-6} 
& wide-angle camera                     & \begin{tabular}[c]{@{}l@{}}eyelid movement\\ from facial images\end{tabular}                                                                     & \begin{tabular}[c]{@{}l@{}}RGB, (eye area) 32$\times$24 pixel,\\ 20 fps\end{tabular}   & \begin{tabular}[c]{@{}l@{}}8 participants: \\ 1068 images for training, 337 images for test
  \end{tabular}   & \cite{mandal2016towards}                             \\ \cline{2-6} 
& Logitech c920 Webcam       & \begin{tabular}[c]{@{}l@{}}head posture \& eye gaze\\ from facial images\end{tabular}                                                     & \begin{tabular}[c]{@{}l@{}}RGB, 1280$\times$720 pixel,\\ 30 fps\end{tabular}           & \begin{tabular}[c]{@{}l@{}} 15 valid participants:\\
398 samples for training, 170 for test\end{tabular}                                           & \cite{vicente2015driver}                             \\ \cline{2-6}  
& \begin{tabular}[c]{@{}l@{}}Allied Vision Tech Guppy \\ Pro F-125 B/C\end{tabular}            & \begin{tabular}[c]{@{}l@{}}eye gaze\\ from facial images\end{tabular}                        & \begin{tabular}[c]{@{}l@{}}grayscale,  800$\times$600 pixel,\\ 30 fps\end{tabular}  & \begin{tabular}[c]{@{}l@{}} 1,860,761 images from 50 participants: \\ 6 classes, unbalanced \end{tabular} & \cite{fridman2015driver}         \\ \cline{2-6}  & IDS UI-3241LE     & image of driver    & IR, 1280$\times$1024 pixel, 30 fps  & 15 participants, totally 12h of video & \cite{martindrive}                             \\ \cline{2-6} & \multirow{3}{*}{Microsoft Kinect}   & image of driver                             & \begin{tabular}[c]{@{}l@{}}RGB, 950$\times$540 pixel, 15 fps;\\ IR, 512$\times$424 pixel, 30 fps;\\ Depth, 512$\times$424 pixel, 30 fps;\end{tabular}  & 15 participants, totally 12h of video & \cite{martindrive}   \\ \cline{3-6} &    & \begin{tabular}[c]{@{}l@{}}head posture \\ from facial images\end{tabular}      & \begin{tabular}[c]{@{}l@{}}RGB, 1920$\times$1080 pixel, 30 fps;\\ Depth, 512$\times$424 pixel, 30 fps;\end{tabular}         &  \begin{tabular}[c]{@{}l@{}}110 sequences: 22 subjects, each 5 recordings \end{tabular}
          & \cite{borghi2017poseidon}  \\ \cline{3-6} &
& \begin{tabular}[c]{@{}l@{}}image of driver \end{tabular}      & \begin{tabular}[c]{@{}l@{}}RGB, 640$\times$360 pixel, 25 fps\end{tabular}  &  \begin{tabular}[c]{@{}l@{}} about 34 thousands images: 10 subjects, each 10-20 min recording \end{tabular}
&\cite{xing2019driver activity} \\ \cline{2-6}& GARMIN VirbX camera  & image of road        & RGB, 1920$\times$1080 pixel, 25 fps    & \begin{tabular}[c]{@{}l@{}} DR(eye)VE dataset: 8 participants, 555,000
frames (74*5-min sequences) \end{tabular} & \cite{palazzi2018predicting}                             \\ \cline{2-6} & unknown camera     & image of driver                      & RGB, 640$\times$480 pixel, 30 fps    &  \begin{tabular}[c]{@{}l@{}}12h of driving: 480 video segments \\
(20 drivers*8 driving conditions*3 segments) \end{tabular} & \cite{li2015predicting}                             \\ \cline{2-6}   & unknown camera     & image of road      & RGB, 320$\times$240 pixel, 15 fps     & \begin{tabular}[c]{@{}l@{}}12h of driving: 480 video segments\\
(20 drivers*8 driving conditions*3 segments) \end{tabular} & \cite{li2015predicting}                            \\ \cline{2-6} & \begin{tabular}[c]{@{}l@{}}ASUS ZenPhone (Z00UD)\\ rear camera\end{tabular}      & image of driver               & RGB, 1920$\times$1080 pixel  & 31 participants & \cite{abouelnaga2017real}    \\ \cline{2-6} & TZYX stereo camera   & image of road  & 512$\times$320 pixel, 32 fps   & \begin{tabular}[c]{@{}l@{}} 139 lane changes: 70 for training, 69 for test  \end{tabular} & \cite{kumar2013learning}                           \\ \cline{2-6}   & unknown camera  & image of driver                                   & 1920$\times$1088 pixel, 25 fps    & Brain4cars: 10 participants, totally 1180 miles of driving & \cite{jain2016brain4cars}                             \\ \cline{2-6}  & unknown camera    & image of road                                 & 720$\times$480 pixel, 30 fps    & Brain4cars: 10 participants, totally 1180 miles of driving & \cite{jain2016brain4cars}                               \\ \cline{2-6}    & Pico Flexx ToF camera   & hand gesture                       & IR, 224$\times$171 pixel, 45 fps   & & \cite{manganaro2019hand}                            \\ \hline
{\multirow{4}{*}{eye tracker}}     & FaceLAB eye tracking device                          & \begin{tabular}[c]{@{}l@{}}eyelid movement, head posture and gaze: \\ blink duration/frequency,\\ PERCLOS,\\ head 3D position/rotations/,\\ saccade frequency, etc.\end{tabular}    & 60 Hz  &\begin{tabular}[c]{@{}l@{}@{}l@{}l}\cite{de2017detection}: 21 participants, 100-110 min per participant, one sample per minute.\\ training set: validation set: test set = 0.7: 0.15: 0.15 \\\cite{liang2007real}: 10 participants, 6*15-min drives per participant  \\\cite{liu2015driver}: 37 participants, 4-fold cross-validation process
\end{tabular}    & \cite{de2017detection}, \cite{liang2007real}, \cite{liu2015driver}    \\ \cline{2-6} 
\multicolumn{1}{|l|}{}                                                                              & Smarteye Pro   & \begin{tabular}[c]{@{}l@{}}eyelid movement: \\ blink duration/frequency,\\ PERCLOS, pupil diameter\end{tabular}                              & 60 Hz &\begin{tabular}[c]{@{}l@{}} 15 participants, 20 km of driving per participant, \\ 398 samples for training, 170 samples for test\end{tabular}                                         & \cite{wang2016driver}     \\ \cline{2-6} 
\multicolumn{1}{|l|}{}  & SMI ETG 2w eye tracking glasses   & eye gaze                  & 1280$\times$720 pixel, 30 fps   & \begin{tabular}[c]{@{}l@{}} DR(eye)VE dataset: 8 participants, 555,000
frames (74*5-min sequences) \end{tabular} & \cite{palazzi2018predicting}                             \\ \cline{2-6} 
\multicolumn{1}{|l|}{}    & Dikablis professional eye tracker                          & eye gaze                      & 384$\times$288 pixel, 60 Hz  & 81 participants, 3 situations per participant, leave-5-out cross validation & \cite{braunagel2017ready}                             \\ \hline
{\multirow{4}{*}{driving simulator}}   & SCANeR Studio                                        & {\begin{tabular}[c]{@{}l@{}}vehicle dynamics:\\ lateral distance from the closest lave and the center of the car;\\ lateral shift of the vehicle center relative to the lane center;\\ time to lane crossing;\\ steering angle/angle velocity, etc.;\\ vehicle speed; number of direction change/out-the-road\end{tabular}}   &   10 Hz      &  \begin{tabular}[c]{@{}l@{}@{}l@{}l@{}l}In \cite{de2017detection}: 21 participants, 100-110 min per participant, one sample per minute.\\ training set: validation set: test set = 0.7: 0.15: 0.15 \\
In \cite{wang2016driver}, 15 participants, 20 km of driving per participant, \\ 398 samples for training, 170 samples for test
 \\ In \cite{tanshi2019modeling}: 36 participants \end{tabular}  & \cite{de2017detection}, \cite{wang2016driver}, \cite{tanshi2019modeling} (20Hz)                    \\ \cline{2-6} 
\multicolumn{1}{|l|}{}                & TNO PreScan      & \begin{tabular}[c]{@{}l@{}}vehicle dynamics:\\ velocity, RPM, gears, etc.\end{tabular}       &       & 30 participants  & \cite{kim2017takeover}  \\ \hline
{\multirow{2}{*}{CAN-Bus}}    & \multirow{2}{*}{CAN-Bus}   & \multirow{2}{*}{\begin{tabular}[c]{@{}l@{}}vehicle dynamics: velocity, steering wheel angle, brake value, \\ RPM acceleration, blinker status\end{tabular}} & &   12h of driving             & \cite{li2015predicting}      \\ \cline{4-6} 
\multicolumn{1}{|l|}{}            &       &                                             & 100 Hz    & 81 participants, 3 situations per participant, leave-5-out cross validation & \cite{braunagel2017ready}                           \\ \hline
\multirow{2}{*}{\begin{tabular}[c]{@{}c@{}}physiological\\ measurement\end{tabular}}           & \begin{tabular}[c]{@{}l@{}}192 channel digital brain wave\\ measurement system from \\ NEURO Company\end{tabular} & electroencephalogram (EEG)                  & & 9 participants  & \cite{fan2010using}                               \\ \cline{2-6}
 & \begin{tabular}[c]{@{}l@{}}Biopac MP150 system \& \\ Acqknowledge software\end{tabular}                           & \begin{tabular}[c]{@{}l@{}}physiological signals:\\ heart rate, sympathetic ratio, vagal ratio, \\ sympathetic-vagal ratio, respiration rate, etc.\end{tabular}   & 1000 Hz  & \begin{tabular}[c]{@{}l@{}} 21 participants, 100-110 min per participant, one sample per minute.\\ training set: validation set: test set = 0.7: 0.15: 0.15 \end{tabular}& \cite{de2017detection}                             \\ \hline
{laser sensor}    & laserBird laser scanner     & head posture                          & 50 Hz  & 81 participants, 3 situations per participant, leave-5-out cross validation & \cite{braunagel2017ready}                             \\ \hline
\multirow{2}{*}{radar \& Lidar sensor}    & unknown radar / Lidar  & relative distance, speed and angle to the surrounding vehicles    & 30 Hz, 360$^{\circ}$ coverage, up to 50 meters                                  & 3 participants, totally 97 drives, average duration of 51 min & \cite{butakov2014personalized}    \\ \cline{2-6}                                    & \begin{tabular}[c]{@{}l@{}}frequency modulated continuous wave\\ mm-wavelength radar\end{tabular}  & hand gesture       & 60 Hz, 0.3mm resolution  & \begin{tabular}[c]{@{}l@{}}6 classes, for training: 5 participants, 20 recordings per gesture\\ for test: 3 participants, 30 recordings per gesture\\ \end{tabular} & \cite{smith2018gesture}                            \\ \hline
{microphone}  & unknown microphone    & speech of driver                                            &   & 12h of driving  & \cite{li2015predicting}                            \\ \hline
\end{tabular}
}

\end{sidewaystable*}
\clearpage
}

\newpage
\onecolumn
\end{document}